\documentclass[10pt,twocolumn,letterpaper]{article}

\usepackage{cvpr}
\usepackage{times}
\usepackage{epsfig}
\usepackage{graphicx}
\usepackage{amsmath}
\usepackage{amssymb}
\usepackage{subfiles}
\usepackage{lipsum}
\usepackage{setspace}

\usepackage[pagebackref=true,breaklinks=true,letterpaper=true,colorlinks,bookmarks=false]{hyperref}

\usepackage{times}
\usepackage{helvet}
\usepackage{courier}
\usepackage{graphicx}
\usepackage{bm}
\usepackage{amsmath,amssymb} 
\usepackage{color}
\usepackage{bbm}
\usepackage{epstopdf}
\usepackage{caption}
\usepackage{subcaption}
\usepackage{enumitem}
\usepackage{calc}
\usepackage{multirow}
\usepackage{xspace}
\usepackage{booktabs}
\usepackage{mathrsfs}
\usepackage{array}
\usepackage[font=small,skip=4pt]{caption}

\newcommand{\figref}[1]{Fig\onedot~\ref{#1}}
\newcommand{\equref}[1]{Eq\onedot~\eqref{#1}}
\newcommand{\secref}[1]{Sec\onedot~\ref{#1}}

\newcommand{\ve}[1]{{\mathbf #1}} 
\newcommand{\hua}[1]{{\mathcal #1}}
\newcommand{\scr}[1]{{\mathcal #1}}

\newcommand{\thickhline}{%
    \noalign {\ifnum 0=`}\fi \hrule height 1pt
    \futurelet \reserved@a \@xhline
}

\newenvironment{myenv}[1]
  {\begin{spacing}{#1}}
  {\end{spacing}}

\DeclareRobustCommand\onedot{\futurelet\@let@token\@onedot}
\def\onedot{\ifx\@let@token.\else.\null\fi\xspace}
\def\eg{\emph{e.g.}} 

\def\any{\forall}
\def\ie{\emph{i.e.}}

\def\etc{\emph{etc}\onedot} 

\def\wrt{w.r.t\onedot} 

\def\etal{\emph{et al.}}

\cvprfinalcopy 

\def\cvprPaperID{2629} 

\ifcvprfinal\fi
\begin{document}

\title{LEGO: Learning Edge with Geometry all at Once by Watching Videos}

\author{
Zhenheng Yang$^{1}$~~Peng Wang$^{2}$~~Yang Wang$^{2}$~~Wei Xu$^{3}$~~Ram Nevatia$^{1}$\\
\\
$^{1}$University of Southern California~~$^{2}$Baidu Research~~\\
$^{3}$National Engineering Laboratory for Deep Learning Technology and Applications\\
}

\maketitle

\begin{abstract}
    Learning to estimate 3D geometry in a single image by watching unlabeled videos via deep convolutional network is attracting significant attention. In this paper, we introduce a ``3D as-smooth-as-possible (3D-ASAP)'' prior inside the pipeline, which enables joint estimation of edges and 3D scene, yielding results with significant improvement in accuracy for fine detailed structures. 
    Specifically, we define the 3D-ASAP prior by requiring that any two points recovered in 3D from an image should lie on an existing planar surface if no other cues provided. We design an unsupervised framework that Learns Edges and Geometry (depth, normal) all at Once (LEGO).
    The predicted edges are embedded into depth and surface normal smoothness terms, where pixels without edges in-between are constrained to satisfy the prior. In our framework, the predicted depths, normals and edges are forced to be consistent all the time.
    We conduct experiments on KITTI to evaluate our estimated geometry and CityScapes to perform edge evaluation. We show that in all of the tasks, \ie depth, normal and edge, our algorithm vastly outperforms other state-of-the-art (SOTA) algorithms, demonstrating the benefits of our approach.
\end{abstract}

\vspace{-0.6\baselineskip}
\section{Introduction}
\vspace{-0.4\baselineskip}
\label{sec:intro}

Humans are highly competent in recovering the 3D geometry of observed natural scenes at very detailed level, even from a single image. 
Practically, being able to do detailed reconstruction for monocular images can be widely applied to many real-world applications such as augmented reality and robotics.

\begin{figure}
\vspace{-0.5\baselineskip}
\includegraphics[width=0.48\textwidth]{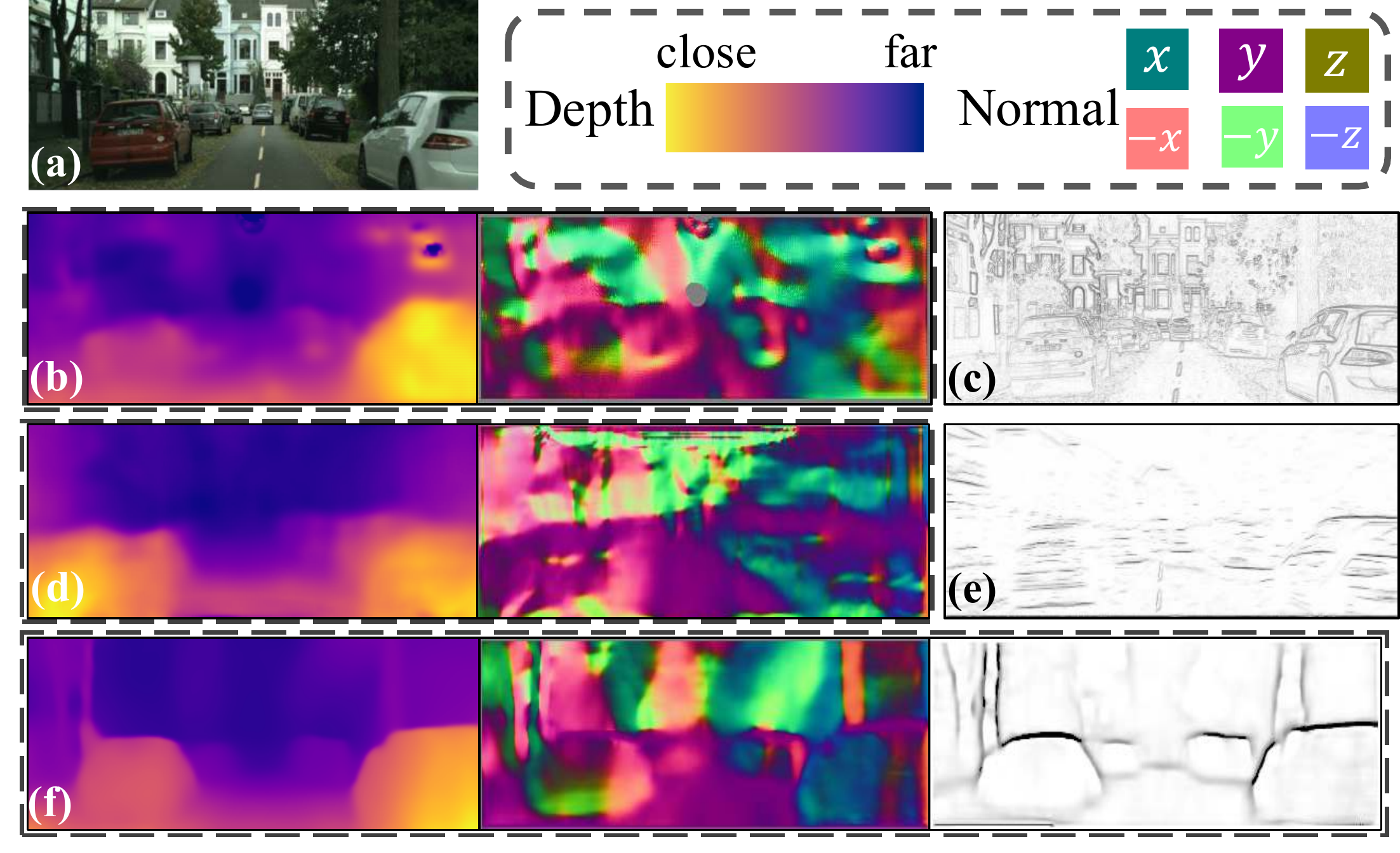}
\caption{(a) Input image; (b) Depth and normal results by \cite{zhou2017unsupervised}; (c) Edges from image gradient; (d) Depth and normal results by \cite{yang2018aaai}; (e) Unsupervised edge detection results by \cite{li2016unsupervised}; (f) Our unsupervised joint estimation of depth, normal and edge.}
\vspace{-0.8\baselineskip}
\label{fig:example}
\end{figure}

Recently, impressive progress~\cite{godard2016unsupervised,zhou2017unsupervised,yang2018aaai} has been made to mimic detailed 3D reconstruction by training a deep network taking only unlabeled videos or stereo images as input and testing on monocular image, yielding even better depth estimation results than those of supervised methods~\cite{eigen2014depth} in outdoor scenarios.
The core underlying idea is the supervision by view synthesis, where the frame of one view (source) is warped to another (target) based on the predicted depths and relative motions, and the photometric error between the warped and observed target frame is used to supervise the training. However, upon a closer look at the predicted results from~\cite{zhou2017unsupervised} in \figref{fig:example}(b) and~\cite{yang2018aaai} in \figref{fig:example}(d), the estimated depths and normals (left and middle) are blurry and do not conform well to the scene geometry.

We argue that this is because the unsupervised learning pipelines are mostly optimizing the per-pixel photometric errors, while paying less attention to the geometrical edges. We use the term ``geometrical edge'' to include depth discontinuities and surface normal changes.
This motivates us to jointly learn an edge representation with the geometry inside the pipeline, so that the two information reinforce each other. We come up with a framework that Learn Edge and Geometry all at Once (LEGO) with unsupversied learning. In our work, 3D geometry helps the model to discover mid-level edges by filtering out the internal edges inside the same surface (those from image gradient as shown in \figref{fig:example}(c)). Conversely, the discovered edges can help the geometry estimation obtain long-range context awareness and non-local regularization, which pushes the model to generate results with fine details.

We formulate the interaction between the two by proposing a ``as smooth as possible in 3D'' (3D-ASAP) prior. It requires all pixels recovered in 3D should 
lay in a same planar surface if no edge exists in-between, such that the edge and the geometrical smoothness are adversarial inside the learning pipeline, yielding consistent and visually satisfying results. 

As shown in \figref{fig:example}(f), the estimated depths and normals of LEGO have consistent structure with the 3D geometry. Compared to the results of SOTA unsupervised edge detection method \cite{li2016unsupervised} in \figref{fig:example}(e), edge results generated by LEGO align well with the scene layout with fewer noises.
The edges discovered in our pipeline is not necessarily semantic but geometrical, arguably alleviating the issues of confusing definition for supervised semantic edge predictions~\cite{MartinFTM01} that was questioned in~\cite{hou2013boundary}. 

We conducted extensive experiments over the public KITTI 2015~\cite{geiger2012we}, CityScapes~\cite{Cordts2016Cityscapes} and Make3D~\cite{saxena2006learning} datasets, and show that LEGO performs much better in depth prediction, especially when transferring the model cross different datasets (relatively 30$\%$ improvements over other SOTA methods). Additionally, LEGO achieves 20$\%$ improvement on normal estimation compared with~\cite{godard2016unsupervised}, and 15$\%$ improvement on geometrical edges detection compared to previous unsupervised edge learning method~\cite{li2016unsupervised}. Lastly, LEGO runs efficiently without much extra computation compared to \cite{zhou2017unsupervised,yang2018aaai}. These demonstrate the efficiency and effectiveness of our approach. We plan to release our code upon the pulication of this paper.

\vspace{-0.5\baselineskip}
\section{Related Work}
\vspace{-0.5\baselineskip}
\label{sec:related}
In this section, we briefly overview some traditional methods, and introduce current SOTA methods for unsupervised single view 3D geometry recovery and edge detection. 

\textbf{~Structure from motion and single view geometry.}
Geometric based methods estimate 3D from a given video with feature matching, such as SFM~\cite{wu2011visualsfm}, SLAM~\cite{mur2015orb,engel2014lsd} and DTAM~\cite{NewcombeLD11}, which could be effective and efficient in many cases. 
However, they can fail at where there is low texture, or drastic change of visual perspective \etc. More importantly, it can not extend to single view reconstruction.
Specific rules are developed for single view geometry, such as computing vanishing point~\cite{HoiemEH07}, following rules of BRDF~\cite{prados2006shape,kong2015intrinsic}, or extract the scene layout with major plane and box representations~\cite{DBLP:conf/iccv/SchwingFPU13,DBLP:conf/3dim/SrajerSPP14} \etc. These methods can only obtain sparse geometry representations, and some of them require certain assumptions (\textit{e.g.} Lambertian, Manhattan world).

\textbf{Supervised single view geometry via CNN.}
Deep neural networks (DCN) developed in recent years, \eg VGG~\cite{simonyan2014very} and ResNet~\cite{laina2016deeper}, provide strong feature representation. Dense geometry, i.e., pixel-wise depth and normal maps, can be readily estimated from a single image~\cite{wang2015designing,eigen2015predicting,laina2016deeper,li2017two,chuang2018cvpr}. The learned CNN model shows significant improvement compared to other methods based on hand-crafted features~\cite{karsch2014depth,ladicky2014pulling,zeisl2014discriminatively}. Others tried to improve the estimation further by appending a conditional random field (CRF)~\cite{DBLP:conf/cvpr/WangSLCPY15,Liu_2015_CVPR,li2015depth}. Recently, Wang \etal \cite{peng2016depth} proposed a depth-normal regularization over large planar surfaces, which is formulated based on a dense CRF~\cite{DBLP:journals/corr/abs-1210-5644}, yielding better results on both depth and normal predictions. However, all these methods require densely labeled ground truths, which are expensive to obtain in natural environments.

\textbf{Unsupervised single view geometry.}
Motivated by traditional methods, videos, which are easier to obtain and hold richer 3D information. Motivated by traditional methods like SFM and DTAM, lots of CNN based methods are proposed to do single view geometry estimation with supervision from vieos, and yield impressive progress. 
 Deep3D \cite{xie2016deep3d} learns to generate the right view from the given left view by supervision of stereo image pairs. In order to do back-propagation on depth values, the depth space is quantized and it is trained to select the right depth value. 
Concurrently, Garg \etal \cite{GargBR16} applied the similar supervision from stereo pairs, while the depth is kept continuous. They apply Taylor expansion to approximate the gradient for depth. Godard \etal \cite{godard2016unsupervised} extend Garg's work by including depth smoothness loss and left-right depth consistency. 
Zhou \etal \cite{zhou2017unsupervised} incoporated camera pose estimation into the training pipeline, which made depth learning possible from monocular videos. And they came up with an explainability mask to relieve the problem of moving object in rigid scenes.
At the same time, Vijayanarasimhan \etal \cite{Vijayanarasimhan17} proposed a network to include the modeling of rigid object motion. Most recently, Yang \etal \cite{yang2018aaai} further induce normal representation, and proposed a dense depth-normal consistency within the pipeline, which not only better regularizes the predicted depths, but also learns to produce a normal estimation. However, as discussed in \secref{sec:intro}, the regularization is only applied locally and can be blocked by image gradient, yielding false geometrical discontinuities inside a smooth surface.

\textbf{Non-local smoothness.} Long range and non-local spatial regularization has been vastly explored in classical graphical models like CRF~\cite{lafferty2001conditional}, where nodes beyond the neighboring are connected, and the smoothness in-between are learned with high-order CRF~\cite{ye2009conditional} or densely-connected CRF~\cite{DBLP:conf/icml/KraehenbuehlK13}. They show superior performance in detail recovery than those with local connections in multiple tasks, \eg segmentation~\cite{DBLP:journals/corr/abs-1210-5644}, image disparity~\cite{scharstein2007learning} and image matting~\cite{chen2013image} \etc In addition, efficient solvers are also developed such as fast bilateral filter~\cite{barron2016fast} or permutohedral lattice~\cite{adams2010fast}. 

Although these methods run effectively and could combine with CNN as a post processing component~\cite{arnab2016higher,crfasrnn_iccv2015,peng2016depth,wang2017occlusion}, they are not very efficient in learning and inference when combined with CNN, due to the iterative loop. To some extent, the non-local information from CRF overlaps with those multi-scale strategies~\cite{zhao2016pyramid,ChenPSA17} proposed recently, which yield comparable performance while are more effective. Thus, we adopt the latter strategy to learn the non-local smoothness inside the unsupervised pipeline, which is represented by geometrical edge in our case.

\textbf{Edge detection.}
Learning edges from an image beyond low level methods such as Sobel or Canny~\cite{canny1986computational} has long been explored via supervised learning~\cite{yamaguchi2012continuous,konishi2003statistical,arbelaez2011contour,DollarICCV13edges} along with the growth of semantic edge datasets~\cite{MartinFTM01,hou2013boundary}. Recently, methods~\cite{bertasius2015high,xie2015holistically,DBLP:journals/corr/Kokkinos15} have achieved outstanding performance by adopting supervisedly trained deep features. 

As discussed, high-level edges can also be learned through non-local smoothness by implicit supervision. One recent work close to ours is \cite{chen2016semantic}. They append a spatial domain transfer (DT) component after a CNN, which acts similar to a CRF for smoothness, and improves the results of semantic segmentation. However, their work is fully supervised with ground truth, and similar to CRF, the DT propagates to neighboring pixels every iteration which is not efficient. When no supervision is provided, Li \etal \cite{li2016unsupervised} proposed to use optical flow~\cite{revaud2015epicflow} to explicitly capture motion edge and use it as supervision for edge models. 

Our method discovers geometrical edges in an unsupervised manner. In addition, we show that it is possible for the network to directly extract edge and smoothen the 3D geometry by enforcing a unified regularization, without appending extra components like~\cite{chen2016semantic}. We also show better performance than~\cite{li2016unsupervised} in street-view cases. 

\vspace{-0.4\baselineskip}
\section{Preliminaries}
\vspace{-0.4\baselineskip}
\label{sec:preliminaries}
In order to make the paper self-contained, we first introduce the preliminaries for unsupervised depth and normal estimation proposed in \cite{zhou2017unsupervised,yang2018aaai}. The core underlying idea is inverse warping from target view to source views with awareness of 3D geometry, and a depth-normal consistency, which we will elaborate in the following paragraphs. 

\textbf{View synthesis as depth supervision.} From the multiple view geometry, we know that for a target view image $I_t$ and a source view image $I_s$, given an estimated depth map $D_t$ for $I_t$ and an estimated transformation $T_{t\rightarrow s} \in \hua{S}\hua{E}(3)$ from $I_t$ to $I_s$, for any pixel $p_t$ in $I_t$, the corresponding pixel $p_s$ in $I_s$ can be found through perspective projection, \ie $p_s \sim \pi(p_t)$. Then, given such a matching relationship, a synthesized target view $\hat{I_s}$ can be generated from $I_s$ through bilinear interpolation. Finally, by comparing the photometric error between the original target view $I_t$ and the synthesized one $\hat{I_s}$. We can supervise the prediction of $D_t$ and $T_{t\rightarrow s}$.
\vspace{-0.1\baselineskip}
Formally, given multiple source views $\hua{S}=\{I_s\}$ from a video sequences close to $I_t$, the photometric loss \wrt $I_t$ can be formulated as,
{\small
\begin{align}
\label{eqn:photometric}
& \scr{L}_{vs}(D, \hua{T})\! =\! \sum\nolimits_{s}\sum\nolimits_{p_t\in I_t}\!|I_t(p_t) - \hat{I_s}(p_t)|, ~\text{s.t.} \any p_t, D(p_t) > 0
\end{align}
}
where $\hua{T}$ is the set of transformations between $I_t$ to each of the source views in $\hua{S}$.


\vspace{-0.1\baselineskip}
\textbf{Regularization of depth.} Nevertheless, supervision based solely on view synthesis is ambiguous, due to one pixel can match to many candidates.
Thus, extra regularization is required to learn reasonable depth prediction. 
One common strategy proposed by previous works \cite{godard2016unsupervised,zhou2017unsupervised} is to encourage the estimated depth to be locally similar when no significant image gradient exists. 
For instance, in \cite{godard2016unsupervised}, the regularization of depth $D_t$ is formulated as: 
{\small
\begin{align}
\label{eqn:regular}
\scr{L}_{s}(D_t, 2) &= \sum_{p_t}\sum_{d \in {x, y}}\|\nabla^{2}_dD_t(p_t)\|_1e^{-\alpha|\nabla_dI(p_t)|}
\end{align}
}
where $\scr{L}_{s}(D, 2)$ is a spatial smoothness term that penalizes L1 norm of second-order gradients of depth along both $x$ and $y$ directions in 2D space. Here, the number $2$ represents the 2nd order. 


\textbf{Regularization with depth-normal consistency.} 
Yang \etal~\cite{yang2018aaai} claim that the smoothness in \equref{eqn:regular} is still a too weak constrain to generate a good scene structure, especially when visualized under normal representation, as shown in 
\figref{fig:example}(b), the predicted normals from~\cite{zhou2017unsupervised} varies on the surface of the ground. In their work, they further introduce a normal map $N_t$ for $I_t$, and a depth-normal consistency energy between $D_t$ and $I_t$ is proposed,
{\small
\begin{align}
\label{eq:orthognal}
&\scr{C}_{p_t}(D_t, N_t) = \sum\nolimits_{p_j\in\hua{N}(p_t)}\omega_{jt}\|(\phi(p_j) - \phi(p_t))^T  N(p_t)||^2 \nonumber
\end{align}
}
where $\scr{N}(p_t)$ is a set of 8-neighbors of $p_t$. $\phi(p)$ is the back projected 3D point from 2D coordinate $p$. $\phi(p_j) - \phi(p_t)$ is a difference vector in 3D, and $\omega_{jt}$ weights the equation.
Based on such an energy, they developed a differentiable depth-to-normal layer to estimate $N_t$ given $D_t$, and a normal-to-depth layer to re-estimate $D_t$ given $N_t$. By applying losses in \equref{eqn:photometric} and \equref{eqn:regular}, plus a first-order normal smootheness loss $\scr{L}_s(N_t, 1)$, $N_t$ can be supervised and $D_t$ can be better regularized with at least 8-neighbors. As shown in~\figref{fig:example}(d), their strategy yields better predicted depths and normals especially along surface regions. The depth and normal consistency same as in \cite{yang2018aaai} is incorporated into LEGO.
\vspace{-0.4\baselineskip}
\section{Learning edge with geometry from videos}
\vspace{-0.4\baselineskip}
\label{sec:approach}
In this section, we introduce the 3D-ASAP prior \wrt geometrical edges, and how the edges can be learned jointly with 3D geometry.

\vspace{-0.2\baselineskip}
\subsection{3D-ASAP prior}
\vspace{-0.2\baselineskip}
Firstly, the core assumption for 3D-ASAP is that for any surface in 3D $S \subset \mathbbm{R}^3$, if there is no other cues provided visually, such as edges, $S$ should be a single 3D planar surface. 
This prior is restrictive for large non-planar surface, but it fits well for street scene which we are mainly dealing with, where the dominant surfaces such as roads, building walls, are still planar.
Formally, it should satisfy the following two conditions,
{\small
\begin{equation}
\ve{\beta}\ve{x}_i + (1-\ve{\beta})\ve{x}_j  \in S \text{~~} \any \ve{x}_i, \ve{x}_j \in S, \beta \in [0, 1]
\label{eqn:asap}
\end{equation}
}
which means any points on the line in-between two points $\ve{x}_i$ and $\ve{x}_j$ should be also inside the surface.
Thus, given a target image $I_t \in \mathbbm{R}^2$, which is a rasterized perspective projection from a set of continuous surfaces $\{S\}$, 
the estimated depth map $D_t$ and normal map $N_t$ should also approximately satisfy such a prior for each $S$. 
Specifically, for $N_t$, any two pixel in the image $p_i$ and $p_j$, we favor the normal of the two points to be the same when $p_i$ and $p_j$ belong to the same $S$, which could be formulated as minimizing,
{\small
\begin{align}
\label{eqn:asap_normal}
\hua{L}_{N} = \sum_{p_i\in I_t}\sum_{p_j\in I_t} \|N_t(p_i) - N_t(p_j)\|_1\kappa(p_i, p_j)
\end{align}
}
where $\kappa(p_i, p_j)$ is a similarity affinity, which is 1 if $p_i, p_j$ in the same $S$, and 0 otherwise.
For $D_t$, we consider a triplet relationship, as indicated in \equref{eqn:asap}. Given two different pixels $p_i$ and $p_j$, we let any pixel $p_k$ on the line in-between, 
lies in the same 3D line with $p_i, p_j$. Formally,
{\small
\begin{align}
\label{eqn:asap_depth}
&\hua{L}_{D} = \sum_{p_i}\sum_{p_j}\!\sum_{p_k\in l(p_i, p_j)}\!\|\ve{g}(p_i, p_j, p_k)\|_1\kappa(p_i, p_k)\kappa(p_j, p_k) \\
&\ve{g}(p_i, p_j, p_k) = \frac{D_t(p_i)-D_t(p_k)}{\phi(p_i)-\phi(p_k)} - \frac{D_t(p_j)-D_t(p_k)}{\phi(p_j)-\phi(p_k)}, \nonumber 
\end{align}
}
where $\phi(p_i) = D_t(p_i)\ve{K}^{-1}h(p_i)$, the back projection function from 2D to 3D space, and $\ve{K}$ is the camera intrinsic and $h(p_i)$ is the homogeneous coordinates of $p_i$. $l(p_i, p_j)$ indicates a set of pixels on the line linking $p_i$ and $p_j$.

\textbf{Approximate with a multi-scale strategy.} If $\kappa()$ is given, such as using image gradient, we can use these two energy functions to serve as non-local smoothness losses for the estimation of depths and normals. 
Nevertheless, it is impractical due to the large number of pixels in an image.
One approximating solution is to drop the dense connection between one pixel with every other pixel to the connection of a set of pixels nearby.
In our case, for each pixel $p_i$, to be compatible with network training, we choose to smoothen normals and depths with its $\hua{N} = {1, 2, 4, 8}$ neighborhood along 3D $x$ and $y$ direction, yielding 16 neighbor pixels, which we found to be sufficiently good to avoid local context. Formally, let $p_i(x, y)$ be the pixel has an offset of $(x, y)$ \wrt $p_i$, the energy for $D_t$ and $N_t$ are changed to be,
{\small
\begin{align}
\label{eqn:asap_app}
\hua{L}_{N} &= \sum_{p_i}\sum_{x,y} \|N_t(p_i) - N_t(p_i(x,y))\|_1\kappa(p_i, p_i(x,y)), \\
\hua{L}_{D_x} &= \sum_{p_i}\sum_{x} \|\ve{g}(p_i, x)\|_1\kappa(p_i, p_i(x))\kappa(p_i, p_i(-x)) \nonumber \\
\ve{g}(p_i, x) &= \frac{D_t(p_i(x))-D_t(p_i)}{\phi(p_i(x))-\phi(p_i)} - \frac{D_t(p_i)-D_t(p_i(-x))}{\phi(p_i)-\phi(p_i(-x))}, \nonumber
\end{align}
}
where $\hua{L}_{D_x}$ means the smoothness along $x$ direction, and $p_i(x)$ is short for $p_i(x, 0)$, similar smoothness is also performed along $y$ direction.
\vspace{-0.2\baselineskip}
\subsection{Parameterize and learn the geometrical edge}
\vspace{-0.1\baselineskip} 
Given the energy loss proposed in \equref{eqn:asap_app}, instead of using image gradient~\cite{yang2018aaai,godard2016unsupervised}, we jointly learn $\kappa()$ by estimating an edge map $E_t$ for the target image. We have,
{\small
\begin{align}
\label{eq:kappa}
 \kappa(p_i, p_j) = \exp\{-\max\nolimits_{p_k \in l(p_i, p_j)}E_t(p_k)\},
\end{align}
}
where $l(p_i, p_j)$ is the line between $p_i$ and $p_j$ including the end points. This indicates the intervening contour cue~\cite{shi2000normalized} for measuring the affinity between two pixels. 

\begin{figure}
\vspace{-0.3\baselineskip}
\includegraphics[width=0.48\textwidth]{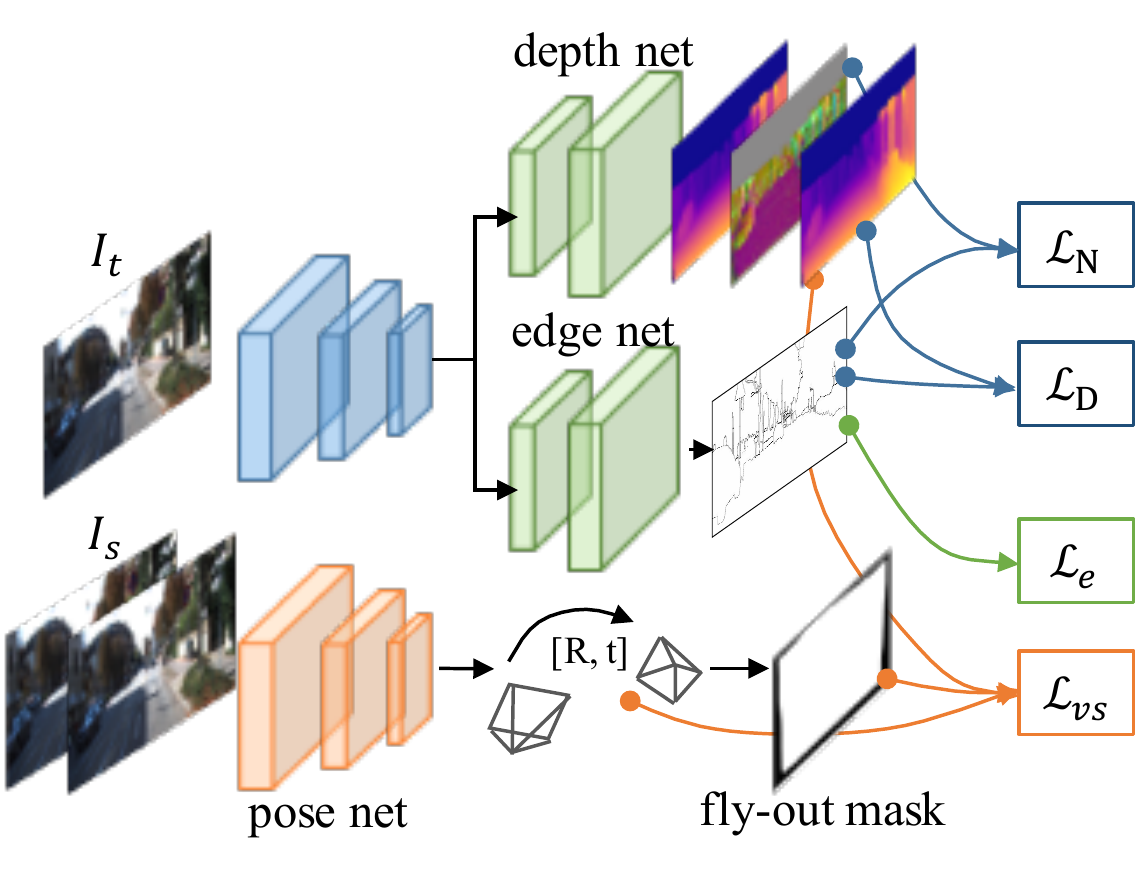}
\caption{Our loss module consists of four parts: visual synthesis loss $\hua{L}_{vs}$, 3D-ASAP losses on depth  and normal maps respectively ($\hua{L}_D$, $\hua{L}_N$), and edge loss $\hua{L}_e$. The same depth-normal consistency as in \cite{yang2018aaai} has been used.}
\vspace{-1.0\baselineskip}
\label{fig:pipe_loss}
\end{figure}

Practically, we parameterize the prediction of $E_t$ using a decoder network, which decodes from a shared image encoder of depth network. Putting \equref{eq:kappa} back into \equref{eqn:asap_app}, plus the photometric losses (\secref{sec:preliminaries}), yields the loss function for both normal map $N_t$, depth map $D_t$ and edge map $E_t$ for regularization. As shown in \figref{fig:pipe_loss}, we show how different components contribute for different losses. 

\textbf{Overcoming the trivial solution.} As we do not have direct supervision for $E_t$, training with \equref{eq:kappa} would result in a trivial solution by predicting every pixel as edge, which perfectly minimize the smoothness both on depths and normals. To resolve this, we add a regularization term with a simple L2 loss to favor no edge predictions, \ie
$\scr{L}_e(E) = \sum_{p_i}||E_t(p_i)||^2$. 
Another potential way is to use cross-entropy as regularization. In our experiment, it does not work well and is very sensitive to the weighting balance. We think it is due to the edge map containing only sparse edges. For supervised learning, HED \cite{xie2015holistically} adopts ground truth to balance positive and negative pixels for the cross-entropy, which is not available in our case.

\begin{figure}
\vspace{-0.3\baselineskip}
\includegraphics[width=0.46\textwidth]{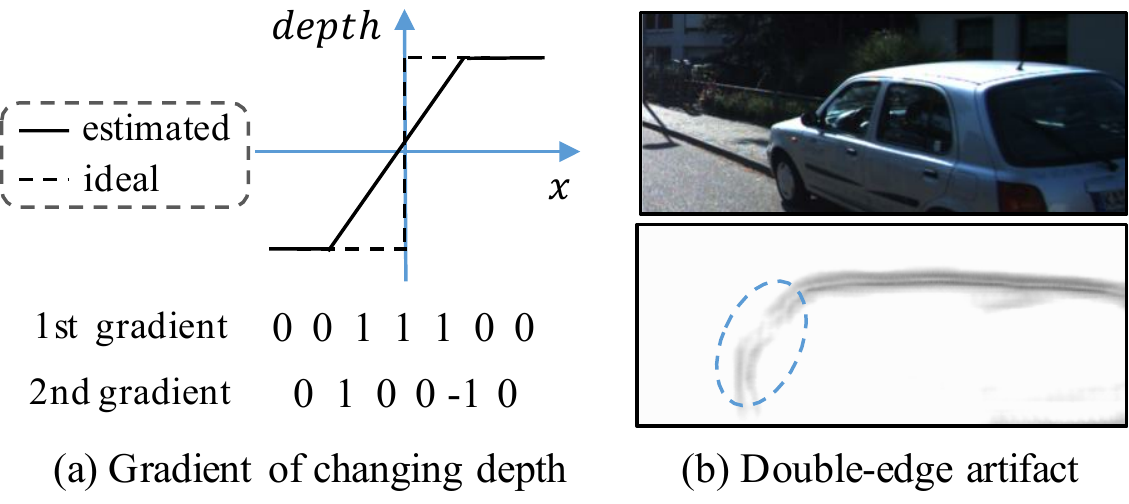}
\caption{Double-edge issue in edge estimation.}
\vspace{-0.5\baselineskip}
\label{fig:double_edge}
\end{figure}

\textbf{Handling double edges during training.} After training using the previous losses, we observe double-edge artifacts, as shown in \figref{fig:double_edge}(b).
Unlike the ideal depth prediction, where depth across a boundary of discontinuity is a step jump (dashed line in \figref{fig:double_edge}(a)), the estimated depth changes smoothly across the object boundary (solid line). Thus, when computing the depth 3D-ASAP regularization term $\hua{L}_{D}$ with one neighborhood in \equref{eqn:asap_app} which is similar to a second-order gradient operation, a non-zero value is generated at both beginning and the end of depth changing. To minimize $\hua{L}_{D}$, the edge map $E_t$ needs to predict a double edge to suppress both of the non-zero values.

We fix this issue by clipping the negative values in the computed gradient map from $\ve{g}(p_i, *)$ in \equref{eqn:asap_app}, as for each boundary along $x$ or $y$ direction, second-order gradient will always have one positive and one negative value. Formally, we replace $\ve{g}(p_i, *)$ to $\ve{g}'=\max(\ve{g}(p_i, *), 0)$.


The architecture of the edge decoder network is set to be the same as the decoder of depth network, while we adopt nearest strategy for edge upsampling from low-scale to high-scale inside the network.
\vspace{-0.5\baselineskip}
\subsection{Overcoming invalid and local gradient}
\begin{figure}
\vspace{-0.3\baselineskip}
\includegraphics[width=0.48\textwidth]{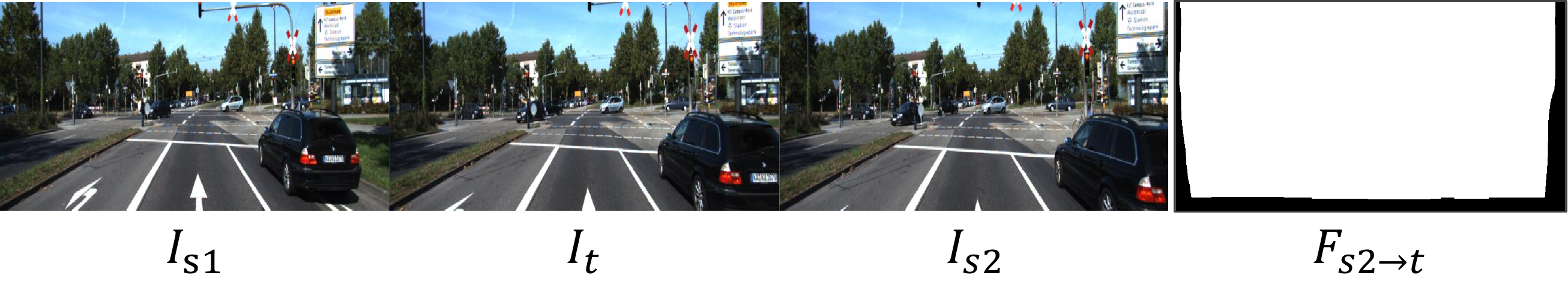}
\caption{Some part of the scene flies out as camera moves forward from $I_t$ to $I_{s2}$. A fly-out mask is calculated from camera motion to filter out such regions.}
\vspace{-0.3\baselineskip}
\label{fig:fly_out}
\vspace{-0.5\baselineskip}
\end{figure}

\vspace{-0.5\baselineskip}
\textbf{Fly-out mask for invalid gradient.} Previous works \cite{zhou2017unsupervised,yang2018aaai} have fixed the length of frame sequence to be 3, with the center frame as the target view ($I_t$) and the neighboring two frames as source view images ($I_{s1}$, $I_{s2}$). When doing view synthesis, possibly part of the corresponding pixels for target view is outside of the source view, yielding invalid gradient for those pixels. As shown in \figref{fig:fly_out}, we identify those pixel and mask out the invalid gradients.


\textbf{Overcoming local gradient.} Similar with gradient locality mentioned in~\cite{zhou2017unsupervised}, the spatial transform operation is based on bilinear interpolation which depends on only 4 neighboring pixels. Thus, loss based on multi-resolution is necessary for effective training. Same strategy is applied in our training pipeline, and in summary, our overall training loss could be written as,
{\small
\begin{align}
&\hua{L}(\{D_l\}, \{N_l\}, \{E_l\}, \hua{T}) =  \sum_l\{\lambda_{vs} \hua{L}_{vs}(D_l, \hua{T}) +\nonumber \\ 
&\text{~~~~~~~~~} \lambda_{d} \hua{L}_{D}(D_l, E_l) + \lambda_{n} \hua{L}_{N}(N_t,E_t)+\lambda_{e} \hua{L}_e(E_t)\}
\end{align}
}
where $\lambda_{vs}, \lambda_{d}, \lambda_{n}, \lambda_{e}$ are balancing factors that are tuned with a sampled validation set from training images.

\begin{figure}
\vspace{-0.4\baselineskip}
\includegraphics[width=0.48\textwidth]{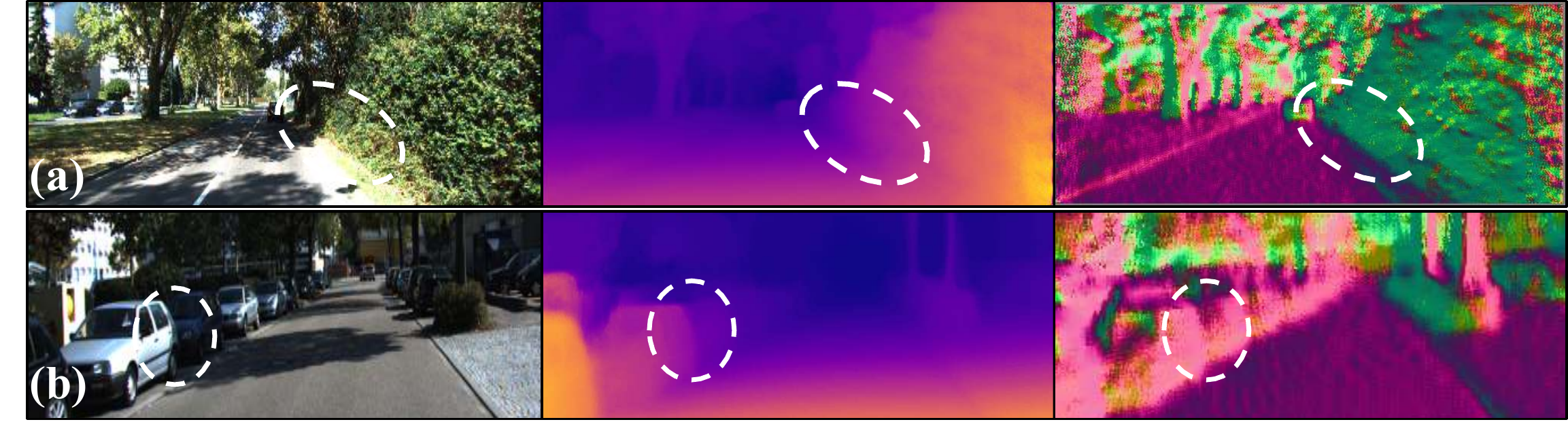}
\caption{Depth and normal are complementary in geometrical edge discovery. (a) Across the edge between two intersecting planes (street and side wall), depth changes smoothly while normal varies drastically; (b) Across the edge between car sides, there is large depth change while normal is uniform.}
\vspace{-0.4\baselineskip}
\label{fig:asap_dn}
\vspace{-0.4\baselineskip}
\end{figure}

Finally, in our experiments, we show it is important to have both smoothness over $D_t$ and $N_t$. As illustrated in Fig. \ref{fig:asap_dn}, depth and normal are complementary for discovering all the geometrical edges. More importantly, the learned edge are consistent with both depth and normal, yielding no perceptual confusion among different information.

\vspace{-0.2\baselineskip}
\section{Evaluation}
\vspace{-0.3\baselineskip}
\label{sec:evaluation}

In this section, we first describe the datasets and evaluation metrics used in our experiments. And then present comprehensive evaluation of LEGO on different tasks. 

\vspace{-0.2\baselineskip}
\subsection{Implementation details}
\vspace{-0.2\baselineskip}
We adopt a DispNet \cite{mayer2016large} like achitecture for depth net and edge net. Regular DispNet is based on an encoder-decoder design with skip connections and multi-scale side outputs. Depth net and edge net share the same encoder while have separate decoder, which decodes depth and edge maps respectively. To avoid artifact grid output from decoder, the kernel size of decoder layers is set to be 4 and the input image is resized to be non-integer times of 64. All \textit{conv} layers are followed by ReLU activation except for the top output layer, where we apply a sigmoid function to constrain the depth and edge prediction within a reasonable range. Batch normalization \cite{ioffe2015batch} is performed on all convolutional layers. To increase the receptive field size while maintaining the number of parameters, dilated convolution with a dilation of 2 is implemented.

During training, Adam optimizer \cite{kingma2014adam} is applied with $\beta_1=0.9$, $\beta_2=0.999$, learning rate of $2\times 10^{-3}$ and batch size of $4$. The balance between different losses is adjusted so that each loss component has loss value of similar scale. In practice, the loss weights are set as: $\lambda_{vs}=1.0$, $\lambda_{d}=2.0$, $\lambda_{n}=0.01$, $\lambda_e=0.15$ for KITTI dataset and $\lambda_{vs}=1.0$, $\lambda_{d}=4.0$, $\lambda_{n}=0.05$, $\lambda_e=0.12$ for Cityscapes dataset. All the hyper-parameters are tuned on the held-out validation set. The input monocular frame sequences are resized to $830 \times 254$ and the length of input sequence is set to be 3. The middle frame serves as the target frame and the neighboring two frames are used as source frames. The whole framework is implemented with Tensorflow platform. On a single Titan X (Pascal) GPU, the framework occupies 4GB of memory with batch size of 4.

\vspace{-0.5\baselineskip}
\subsection{Datasets and metrics}
\vspace{-0.5\baselineskip}
We conducted experiments on different tasks: depth estimation, normal estimation and edge detection. The performances are evaluated on three popular datasets: KITTI 2015, Cityscapes and Make3D, using corresponding metrics.

\textbf{KITTI 2015.}
KITTI 2015 dataset provides videos in 200 street scenes captured by stereo RGB cameras, and sparse depth ground truth captured by Velodyne laser scanner. During training, 156 videos excluding test scenes are used, with the left and right videos treated independently. The training sequences are constructed with three consecutive frames, resulting in 40250 training samples. There are two test splits of KITTI 2015: the official test set consisting of 200 images (KITTI split) and the test split proposed in \cite{eigen2014depth} consisting of 697 images (Eigen split). The official KITTI test split provides ground truth of better quality compared to Eigen split, where less than 5\% pixels in the input image has ground truth depth values. LEGO is evaluated on both splits to better compare with other methods.

\textbf{Cityscapes.} 
Cityscapes is a city-scene dataset with ground truth for semantic segmentation. It contains 27 stereo videos, and provides pixel-wise semantic segmentation ground truth for 500 frames in validation split. Training sequences are constructed from 18 left-view videos of the training set, resulting in 69728 training samples. The semantic segmentation ground truth in 500 validation frames is used for the evaluation of edge detection. Details of using segmentation ground truth are described in Sec. \ref{edge_exp}.

\textbf{Make3D.} Make3D dataset contains no videos but 534 monocular image and depth ground truth pairs. Unstructured outdoor scenes, including bush, trees, residential buildings, \etc are captured in this dataset. Same as in \cite{zhou2017unsupervised,godard2016unsupervised}, the evaluation is performed on the test set of 134 images.

\textbf{Metrics.} The existing metrics of depth, normal and edge detection have been used for evaluation, as in \cite{eigen2014depth}, \cite{fouhey2013data} and \cite{arbelaez2011contour}. For depth and edge evaluation, we have used the code by \cite{godard2016unsupervised} and \cite{DollarICCV13edges} respectively. For normal evaluation, we implement the evaluation metrics in \cite{fouhey2013data} and verify it by validating the results in \cite{eigen2015predicting}. The explanation of each metric used in our evaluation is specified in Tab. \ref{metrics}. 


\begin{table}[!htbp]
\vspace{-0\baselineskip}
\centering
\fontsize{8}{10}\selectfont
\def\arraystretch{1.5}
\caption{From top row to bottom row: depth, normal and edge evaluation metrics.}
\setlength{\tabcolsep}{2pt}
\begin{tabular}{l|l}
\specialrule{.2em}{.1em}{.1em}
Abs Rel: $\!\frac{1}{|D|}\!\sum_{d'\in D}\!|d^*\!\!\!-\!\!d'|/d^*$       & Sq Rel: $\frac{1}{|D|}\!\sum_{d'\in D}\!||d^*\!\!\!-\!\!d'||^2\!/d^*$                \\
RMSE: $\!\sqrt{\!\frac{1}{|D|}\!\sum_{d'\!\in\! D}||d^*\!\!\!-\!\!d'||^2}$    & RMSE log: $\!\sqrt{\!\!\frac{1}{|D|}\!\!\sum_{d'\!\in\! D}\!\!||\!\log\! d^*\!\!\!-\!\!\log\! d'||^2\!}\!$ \\ \hline
mean: $\frac{1}{|N|}\!\sum_{n'\!\in\! N}(n^*\!\!\cdot\! n')$            & median: $median([(n^*\!\!\cdot\! n')]_{n'\!\in\! N})$                            \\
$X^{\circ}$: $\!\%$ \!of $n'\!\!\in \!\!N\!$, $(n^*\!\!\cdot\! n'\!)\!<\!X^{\circ}$   &                                                                      \\ \hline
ODS: optimal F1 for the dataset & OIS: optimal F1 for each image              \\
AP: average precision & PR curve: precision-recall curve \\ \hline
\end{tabular}
\label{metrics}
\vspace{-1.2\baselineskip}
\end{table}

\subsection{Depth and normal experiments}
\textbf{Experiment setup.}
The depth and surface normal experiments are conducted on KITTI 2015, Cityscapes and Make3D datasets. For KITTI 2015, the given depth ground truth is used for evaluting depth estimation, and the normal ground truth is computed from interpolated depth ground truth using depth-to-normal layer. Videos in Cityscapes dataset are captured by the cameras mounted on moving cars. Part of the car is captured in the videos hence the bottom part of the frames is cropped. As no ground truth depth is given in this dataset, we are using Cityscapes only for training. Images in Make3D dataset have different aspect ratio from KITTI or Cityscapes frames, the central part is cropped out for evaluation. For both depth and normal evaluation, only pixels with ground truth depth values are evaluated. One LEGO variant is generated by removing fly-out mask from the pipeline, LEGO (no fly-out), to explore the effectivenss of fly-out mask.

The following evaluations are performed to present the depth and normal results: (1) depth estimation performance compared with SOTA methods; (2) normal estimation performance compared with SOTA methods; (3) generalization capability between different datasets.


\textbf{Comparison with state-of-the-art.}
The model is trained on KITTI 2015 raw videos excluding frames of scenes in both test splits. Following the tradition of other methods \cite{eigen2014depth,zhou2017unsupervised,godard2016unsupervised}, the maximum of depth estimation on KITTI split is capped at 80 meters and the same crop as in \cite{eigen2014depth} is applied during evaluation on Eigen split. 

\begin{figure*}
\vspace{-0.5\baselineskip}
\centering
\includegraphics[width=\textwidth]{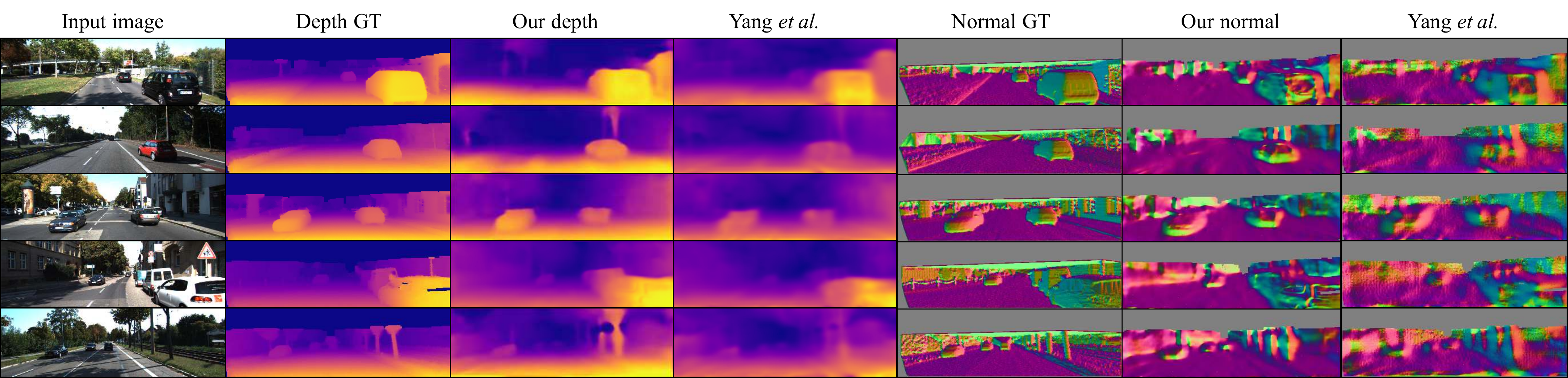}
\caption{Visual comparison between Yang \etal \protect\cite{yang2018aaai} and LEGO results on KITTI test split. The depth and normal ground truths are interpolated and all images are reshaped for better visualization. For depths, LEGO results have noticeably shaper edges and the depth edges are well aligned with object boundaries. For surface normals, LEGO results have fewer artifacts and extract clear scene layout.}
\vspace{-0.5\baselineskip}
\label{fig:examples}
\end{figure*}

Tab. \ref{tbl:sota} shows the comparison of LEGO variants and recent SOTA methods. LEGO outperforms all unsupervised methods \cite{zhou2017unsupervised,kuznietsov2017semi,yang2018aaai} consistently on both test splits and performs comparably to the semi-supervised method \cite{godard2016unsupervised}. It is also worth noting that on the metric of ``Sq Rel'', LEGO outperforms all other methods on KITTI split. This metric measures the ratio of square of prediction error over the ground truth value, and thus is sensitive to points where the depth values are away from the ground truth. The good performance under this metric indicates that LEGO produces consistent 3D scene layout and generates fewer outlier depth values.

\begin{table}[t]
\vspace{-0.2\baselineskip}
\centering
\caption{Monocular depth evaluation results on KITTI split (upper part) and Eigen split(lower part). All methods use KITTI dataset for traning if not specially noted. Results of \cite{zhou2017unsupervised} on KITTI test split are generated by training their released model on KITTI dataset. \textit{CS} denotes the method trained on Cityscapes and then finetuned KITTI data. \textit{PP} and \textit{R} denote post processing and ResNet respectively.}
\label{tbl:sota}
\fontsize{6.5}{7.5}\selectfont
\bgroup
\def\arraystretch{1.2}
\setlength{\tabcolsep}{1.5pt} 
\begin{tabular}{lcccccc}
\specialrule{.2em}{.1em}{.1em}
\multirow{2}{*}{Method}                                      & \multirow{2}{*}{Test data}                        & \multirow{2}{*}{Supervision} & \multicolumn{4}{c}{Lower the better} \\ \cline{4-7} 
                                                             &                                                   &         & Abs Rel  & Sq Rel & RMSE  & RMSE log \\ \hline
\multicolumn{1}{l|}{Train set mean}                          & \multicolumn{1}{l|}{\multirow{7}{*}{KITTI split}}            & Depth   & 0.398    & 5.519  & 8.632 & 0.405    \\
\multicolumn{1}{l|}{Godard \etal\cite{godard2016unsupervised}}                  & \multicolumn{1}{r|}{}                             &Pose     & 0.124    & 1.388  & 6.125 & 0.217    \\
\multicolumn{1}{l|}{Vij.. \etal\cite{Vijayanarasimhan17}}        & \multicolumn{1}{r|}{}                  &     & -        & -      & -     & 0.340     \\
\multicolumn{1}{l|}{Zhou \etal\cite{zhou2017unsupervised}}                    & \multicolumn{1}{l|}{\multirow{2}{*}{}}            &       & 0.216    & 2.255  & 7.422 & 0.299    \\
\multicolumn{1}{l|}{Yang \etal\cite{yang2018aaai}}                    & \multicolumn{1}{l|}{\multirow{2}{*}{}}            &       & 0.165    & 1.360  & 6.641 & 0.267    \\ 
\multicolumn{1}{l|}{LEGO (no fly-out)}                                    & \multicolumn{1}{l|}{}                             &                & 0.157    & 1.303  & 6.223 & 0.241 \\
\multicolumn{1}{l|}{LEGO}                                    & \multicolumn{1}{l|}{}                             &       & 0.154   & 1.272  & 6.012 & 0.230     \\ \hline
\multicolumn{1}{l|}{Godard \textit{et al.}\cite{godard2016unsupervised}+CS}      & \multicolumn{1}{c|}{\multirow{3}{*}{KITTI split}} & Pose                                                 & 0.104 & 1.070 & 5.417 & 0.188 \\
\multicolumn{1}{l|}{Godard \textit{et al.}\cite{godard2016unsupervised}+CS+PP+R}   & \multicolumn{1}{c|}{}                             & Pose                                          & 0.097 & 0.896 & 5.093 & 0.176 \\
\multicolumn{1}{l|}{LEGO+CS}                                 & \multicolumn{1}{c|}{}                             &                                                      & 0.142 & 1.237 & 5.846 & 0.225 \\ \hline
\multicolumn{1}{l|}{Train set mean}                          & \multicolumn{1}{l|}{\multirow{8}{*}{Eigen split}} & Depth       & 0.403    & 5.530  & 8.709 & 0.403     \\
\multicolumn{1}{l|}{Kuz.. \etal\cite{kuznietsov2017semi} supervised}   & \multicolumn{1}{l|}{}                             & Depth        & 0.122    & 0.763  & 4.815 & 0.194    \\
\multicolumn{1}{l|}{Kuz.. \etal\cite{kuznietsov2017semi} unsupervised} & \multicolumn{1}{l|}{}                             &      Pose    & 0.308    & 9.367  & 8.700 & 0.367    \\
\multicolumn{1}{l|}{Kuz.. \textit{et al.}\cite{kuznietsov2017semi} combined} & \multicolumn{1}{l|}{}                      & Pose+Depth                                           & 0.113 & 0.741 & 4.621 & 0.189 \\
\multicolumn{1}{l|}{Godard \etal\cite{godard2016unsupervised}}                  & \multicolumn{1}{l|}{}                             &     Pose     & 0.148    & 1.344  & 5.927 & 0.247    \\
\multicolumn{1}{l|}{Zhou \etal\cite{zhou2017unsupervised}}                    & \multicolumn{1}{l|}{}                             &             & 0.208    & 1.768  & 6.856 & 0.283     \\
\multicolumn{1}{l|}{Yang \etal\cite{yang2018aaai}}                    & \multicolumn{1}{l|}{}                             &             & 0.182    & 1.481  & 6.501 & 0.267     \\
\multicolumn{1}{l|}{LEGO (no fly-out)}                                    & \multicolumn{1}{l|}{}                             &                & 0.170    & 1.382  & 6.321 & 0.255 \\
\multicolumn{1}{l|}{LEGO}                                    & \multicolumn{1}{l|}{}                             &                & 0.162    & 1.352  & 6.276 & 0.252    \\ \hline
\multicolumn{1}{l|}{Godard \textit{et al.}\cite{godard2016unsupervised}+CS}      & \multicolumn{1}{l|}{\multirow{4}{*}{Eigen split}}                                                   & Pose                                                 & 0.124 & 1.076 & 5.311 & 0.219 \\
\multicolumn{1}{l|}{Godard \textit{et al.}\cite{godard2016unsupervised}+CS+PP+R}   & \multicolumn{1}{l|}{}                                                   & Pose                                          & 0.114 & 0.898 & 4.935 & 0.206 \\
\multicolumn{1}{l|}{Zhou \textit{et al.}\cite{zhou2017unsupervised}+CS}        & \multicolumn{1}{l|}{}                                                   &                                                      & 0.198 & 1.836 & 6.565 & 0.275 \\
\multicolumn{1}{l|}{LEGO+CS}                                 &    \multicolumn{1}{l|}{}                                                &                                                      & 0.159 & 1.345 & 6.254 & 0.247 \\ \hline

\end{tabular}
\egroup
\vspace{-1.5\baselineskip}
\end{table}

The normal ground truth is generated by applying depth-to-normal layer on interpolated depth ground truth. As the depth ground truth point in Eigen split is very sparse (\textless5\%), the interpolation incorporates extra noise and not suitable for normal evaluaton. The normal evaluation is performed only on KITTI split. The comparison of normal evaluations on KITTI split is presented in Tab. \ref{tbl:normal}. The methods we have compared with include: (1) ground truth normal mean: mean value of ground truth normal over the image size; (2) pre-defined scene: based on the observation that KITTI is a street scene dataset, the image is divided into 4 parts by connecting the center and 4 corners, approximating the scene with road in the bottom part, buildings on the two sides and sky at the top; (3) normal results generated by applying depth-to-normal layer on depth maps from some baseline methods \cite{godard2016unsupervised,zhou2017unsupervised,yang2018aaai}.

LEGO outperforms all baseline methods by a large margin. Note that LEGO has inferior results compared to \cite{godard2016unsupervised} on depth results while still outperforms on normals. One possible reason is that the depth is only evaluated on pixels with ground truth values, while the normal direction of each pixel is computed based on neighboring points, which indicates that LEGO may produce depth and normal that are more consistent with the scene layout. Compared to LEGO (no fly-out), LEGO experiences larger performance improvement in normal results compared to depth evaluation.

Qualitative results are shown in Fig. \ref{fig:examples}. Compared with \cite{yang2018aaai}, LEGO generates smoother depth and normal outputs within the same surface while still preserving clear geometrical edges.

\begin{table}[t] \small
\vspace{-0.2\baselineskip}
\centering
\caption{Normal evaluation results on KITTI test split.}
\label{tbl:normal}
\fontsize{6.5}{7}\selectfont
\bgroup
\def\arraystretch{1.2}
\begin{tabular}{l|c|c|c|c|c}
\specialrule{.2em}{.1em}{.1em}
Method                        & Mean  & Median & $11.25^{\circ}$ & $22.5^{\circ}$  & $30^{\circ}$    \\ \hline
Ground truth normal mean      & 72.39 & 64.72  & 0.031 & 0.134 & 0.243 \\
Pre-defined scene             & 63.52 & 58.93  & 0.067 & 0.196 & 0.302 \\
Zhou \etal \cite{zhou2017unsupervised} & 50.47 & 39.16  & 0.125 & 0.303 & 0.425 \\
Godard \etal \cite{godard2016unsupervised} & 39.28 & 29.37 & 0.158 & 0.412 & 0.496 \\
Yang \etal \cite{yang2018aaai}                          & 47.52 & 33.98  & 0.149 & 0.369 & 0.473 \\
LEGO (no-flyout)                  & 39.29 & 28.14 & 0.226 & 0.421 & 0.508 \\ 
LEGO                    & \textbf{36.13} & \textbf{25.94} & \textbf{0.241} & \textbf{0.473} & \textbf{0.542} \\ \hline
\end{tabular}
\egroup
\vspace{-0.3\baselineskip}
\end{table}

\begin{table}[t] \small
\vspace{-0.3\baselineskip}
\centering
\caption{Depth evaluation results with model trained on a different dataset. Note that \cite{godard2016unsupervised} leverages pose ground truth during training.}
\label{tbl:generalization}
\fontsize{6.5}{7}\selectfont
\bgroup
\def\arraystretch{1.1}
\setlength{\tabcolsep}{2.5pt}
\begin{tabular}{lccccc}
\specialrule{.2em}{.1em}{.1em}
\multirow{2}{*}{Methods}                          & \multirow{2}{*}{Train/Test dataset}               & \multicolumn{4}{c}{Lower the better} \\ \cline{3-6} 
                                                  &                                                   & Abs Rel  & Sq Rel & RMSE  & RMSE log \\ \hline
\multicolumn{1}{l|}{Godard \etal\cite{godard2016unsupervised}}     & \multicolumn{1}{c|}{\multirow{3}{*}{CS/K}}        & 0.699    & 10.06  & 14.44 & 0.542    \\
\multicolumn{1}{l|}{Zhou \etal\cite{zhou2017unsupervised}}       & \multicolumn{1}{c|}{}                             & 0.275    & 2.883  & 7.684 & 0.382    \\
\multicolumn{1}{l|}{LEGO}                         & \multicolumn{1}{c|}{}                             & \textbf{0.201}    & \textbf{1.650}  & \textbf{6.788} & \textbf{0.278}    \\ \hline
\multicolumn{1}{l|}{Godard \textit{et al.}\cite{godard2016unsupervised}}     & \multicolumn{1}{c|}{\multirow{3}{*}{CS/Make3D}} & 0.535    & 11.99  & 11.51 & -        \\
\multicolumn{1}{l|}{Zhou \textit{et al.}\cite{zhou2017unsupervised}}       & \multicolumn{1}{c|}{}                             & 0.383    & \textbf{5.321}  & 10.47 & 0.478    \\
\multicolumn{1}{l|}{LEGO}                       & \multicolumn{1}{c|}{}                             & \textbf{0.352} & 7.731  & \textbf{7.194}  & 0.346    \\ \hline
\multicolumn{1}{l|}{Kuznietsov \etal\cite{kuznietsov2017semi}} & \multicolumn{1}{c|}{\multirow{1}{*}{K/Make3D}}                             & 0.421    & -      & 8.237 & 0.190    \\ \hline
\end{tabular}
\egroup
\vspace{-0.8\baselineskip}
\end{table}

\begin{figure}
\vspace{-0.3\baselineskip}
\includegraphics[width=0.48\textwidth]{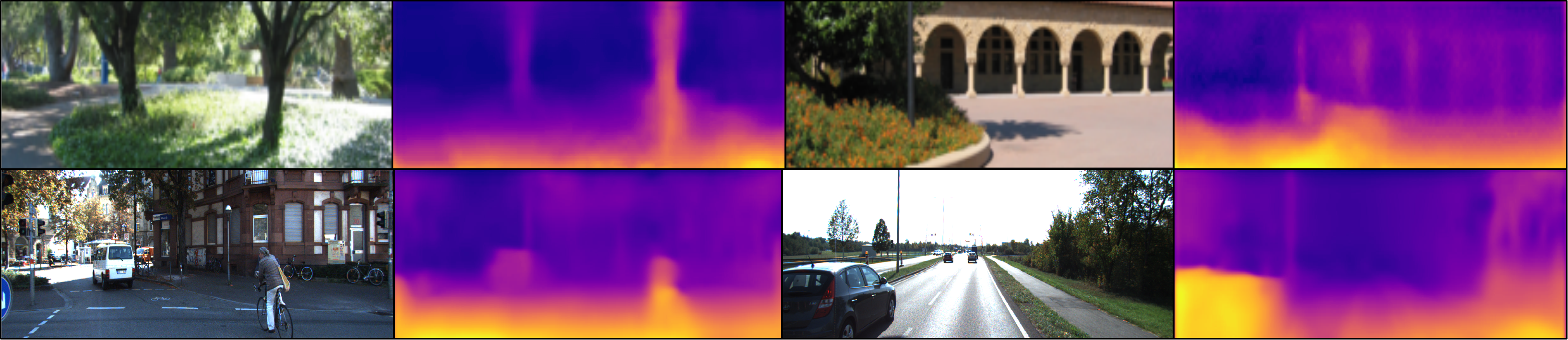}
\caption{Depth test results by the model trained on a different dataset. Top row: trained on Cityscapes, tested on Make3D. Second row: trained on Cityscapes, tested on KITTI 2015.}
\vspace{-1.0\baselineskip}
\label{fig:generalization}
\end{figure}

\textbf{Generalization capability.} Generalizing to data unseen during training is an important property for unsupervised geometry estimation as there may not be enough data for certain scenes. The generalization capability of LEGO is tested by training on one dataset and testing on another dataset. Specifically, to compare with previous methods, two experiments have been conducted: (1) pipeline trained on Cityscapes dataset (CS) and tested on KITTI dataset (K); (2) pipeline trained on Cityscapes and evaluated on Make3D dataset (Make3D). 
The comparison results are shown in Tab. \ref{tbl:generalization}.

Under both settings, LEGO achieves state-of-the-art performance. When transferring from Cityscapes to KITTI, it outperforms other methods by a large margin. One potential explanation is that compared to supervised or semi-supervised methods, LEGO has less risk of overfitting. Compared to other unsupervised methods, our novel 3D-ASAP regularization encourages the network to learn the structural layout information jointly and thus the trained model is more robust to scene changes. Some visualization examples of the generalization results are shown in Fig. \ref{fig:generalization}.

\vspace{-0.15\baselineskip}
\subsection{Edge experiments}
\vspace{-0.15\baselineskip}
\textbf{Experiment setup.} 
\label{edge_exp}
The geometrical edge detection performance is evaluated on Cityscapes dataset. Cityscapes contains a validation set of 500 images with pixel-wise semantic segmentation annotation. The edge ground truth is generated from the segmentation ground truth. Some geometrically connected categories such as ``ground'' and ``road'', ``fence'' and ``guard rail'', ``pole'' and ``traffic sign'' are combined and the geometrical edges are extracted from the boundaries of these combined categories. Fig. \ref{fig:edge_gt} shows how the ground truth edge ground truth is generated. More details are provided in the supplementary material.

\begin{figure}
\vspace{-0.2\baselineskip}
\includegraphics[width=0.48\textwidth]{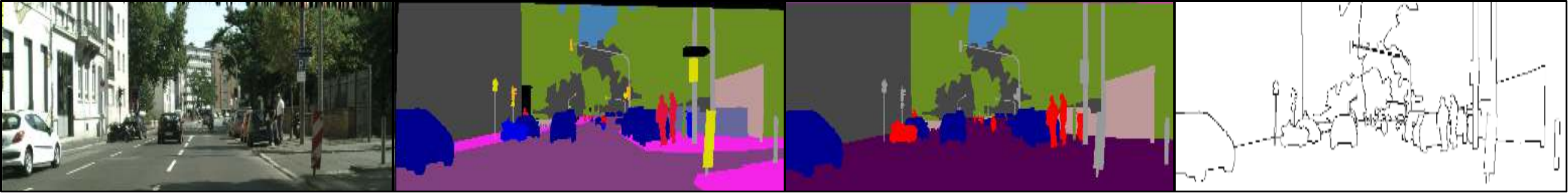}
\caption{Display of the process of geometric edge ground truth generation. From left to right: RGB image, original segmentation ground truth, combined segmentation result, edge ground truth.}
\vspace{-1.0\baselineskip}
\label{fig:edge_gt}
\end{figure}

As there has not been previous work that reported edge detection performance on Cityscapes, we compare with unsupervised edge learning \cite{li2016unsupervised} and some other baselines we build.
The results of \cite{li2016unsupervised} are generated by traning their public model on Cityscapes videos. Different from \cite{li2016unsupervised} which randomly samples training data, we do not apply any sampling to make the number of training samples comparable to our method. Other baseline methods include: (1) modification of Zhou \etal \cite{zhou2017unsupervised} method by adding an edge detection network to the model (Zhou \etal \cite{zhou2017unsupervised}+edge net); (2) apply the pre-trained Structured Edge detector (SE) \cite{DollarICCV13edges} on depth and normal output from \cite{yang2018aaai} (SE-D/SE-N); (3) apply the pre-trained holistically-nested edge detector (HED) \cite{xie2015holistically} edge detector on depth and normal results from \cite{yang2018aaai} (HED-D/HED-N).

\textbf{Ablation study.} Two LEGO variants are generated by applying geometrical edge in only depth or normal smoothness term (LEGO (d-edge) and LEGO (n-edge)). We explore the effect of depth and normal complementing each other in geometrical edge detection.

\begin{figure}
\vspace{-0.2\baselineskip}
\includegraphics[width=0.5\textwidth]{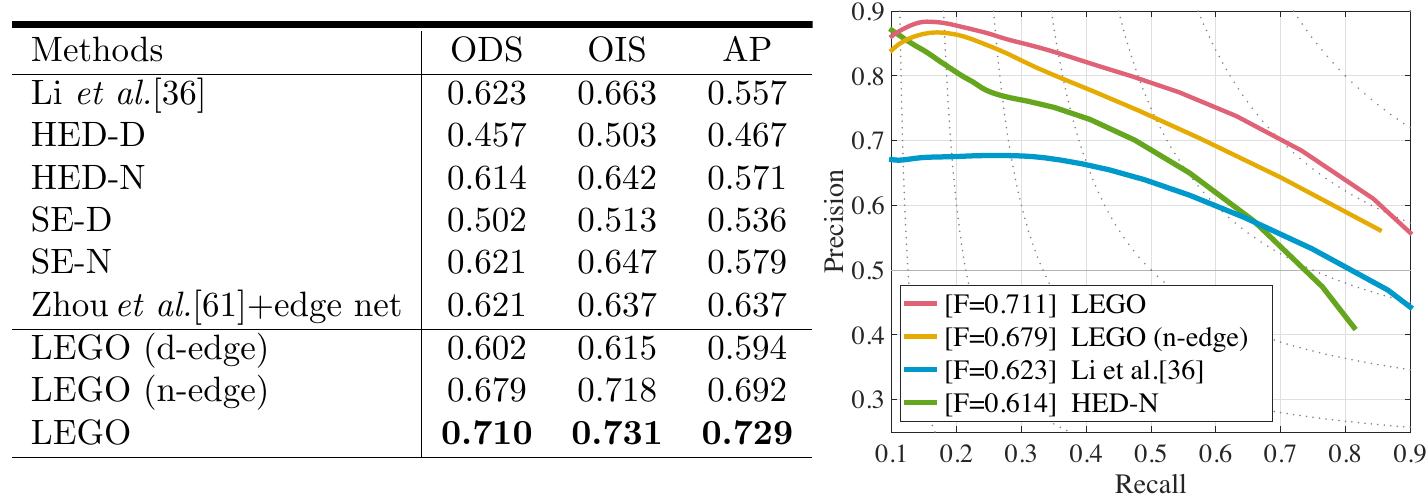}
\caption{Edge evaluation results on Cityscapes.}
\vspace{-0.3\baselineskip}
\label{fig:edge_eval}
\end{figure}

\begin{figure}
\vspace{-0.3\baselineskip}
\includegraphics[width=0.5\textwidth]{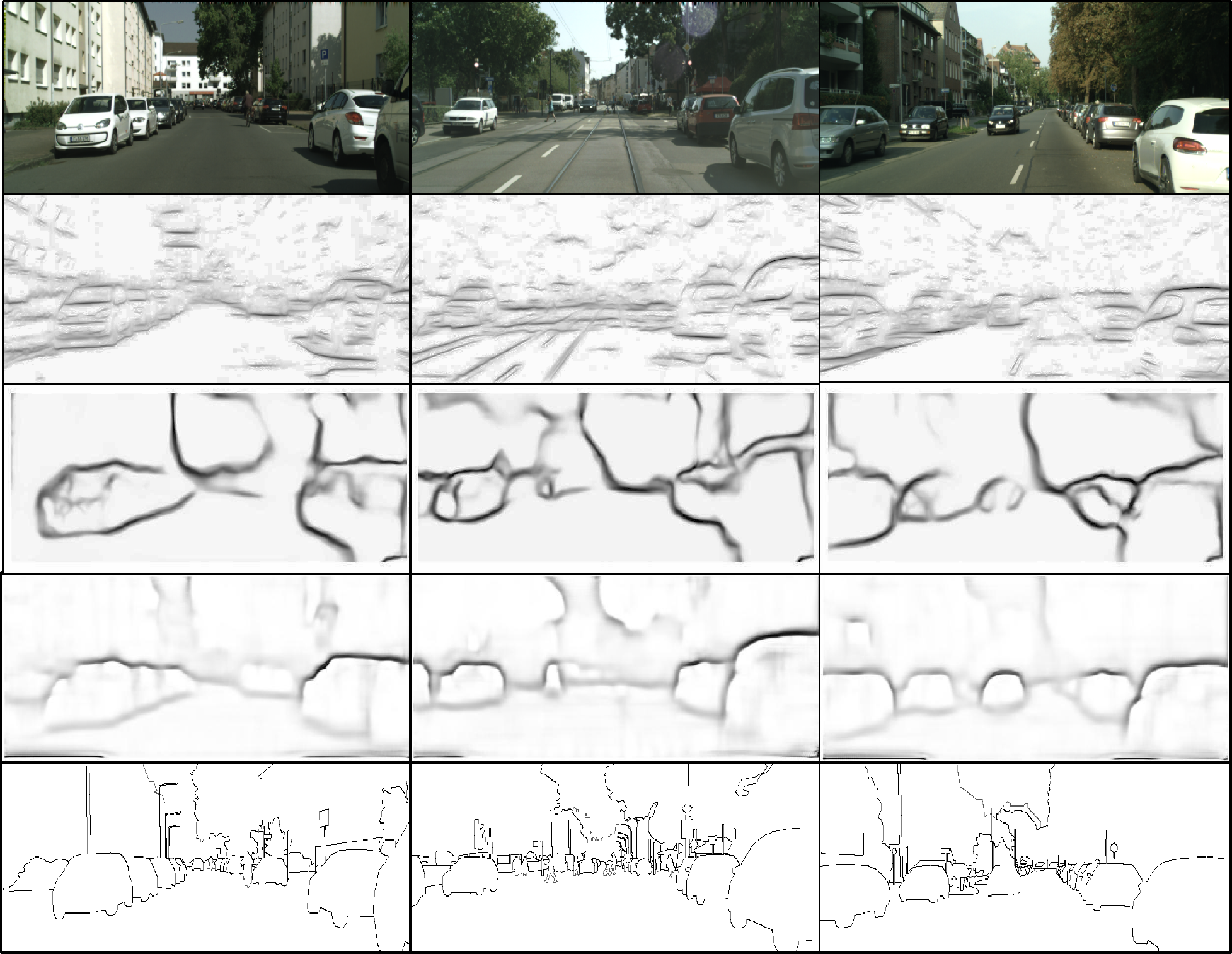}
\caption{Edge detection results on Cityscapes dataset. From top to bottom: input image, unsupervised edge by Li \etal \protect\cite{li2016unsupervised}, HED-N, our results, edge ground truth. All detection visualization results are before the process of non-maximum suppression).}
\vspace{-0.8\baselineskip}
\label{fig:edge_example}
\end{figure}

\textbf{Comparison with other methods.} LEGO is compared with re-trained \cite{li2016unsupervised} and general edge detection (SE \cite{DollarICCV13edges}, HED \cite{xie2015holistically}) results applied on depth/normal output. The quantitative and qualitative results are presented in Fig. \ref{fig:edge_eval} and Fig. \ref{fig:edge_example}. LEGO outperforms other methods by a large margin on all metrics. In visualization results as in \figref{fig:edge_example}, predictions by LEGO preserve the object boundaries and ignore trivial edges within a surface like lane marking. Compared to the edge generated from normal (HED-N), LEGO estimations are well aligned with ground truth edges.

\vspace{-0.5\baselineskip}
\section{Conclusion} 
\vspace{-0.3\baselineskip}
In this paper, we proposed LEGO, an unsupervised framework for joint depth, normal and edge learning. A novel 3D-ASAP prior is proposed to better regularize the learning of scene layout. This regularization jointly considers the three important descriptors of 3D scene and improves the results on all tasks: depth, normal and edge estimation.
We conducted comprehensive experiments to present the performance of LEGO. On KITTI dataset, LEGO achieves SOTA performance on both depth and normal evaluation. 
For edge evaluation, LEGO outperformes the other methods by a large margin on Cityscapes dataset. 


\setcounter{section}{0}
\setcounter{figure}{0}
\setcounter{table}{0}
\onecolumn
\begin{myenv}{1.6}
\begin{center}
\textbf{{\fontsize{14}{16}\selectfont LEGO: Learning Edge with Geometry all at Once by Watching Videos\\
Suppelmentary Material}}
\end{center}
\end{myenv}
\maketitle

\vspace{1.2\baselineskip}
\section{Edge ground truth generation}

\begin{figure*}[!htbp]
\vspace{\baselineskip}
\centering
\includegraphics[width=\textwidth]{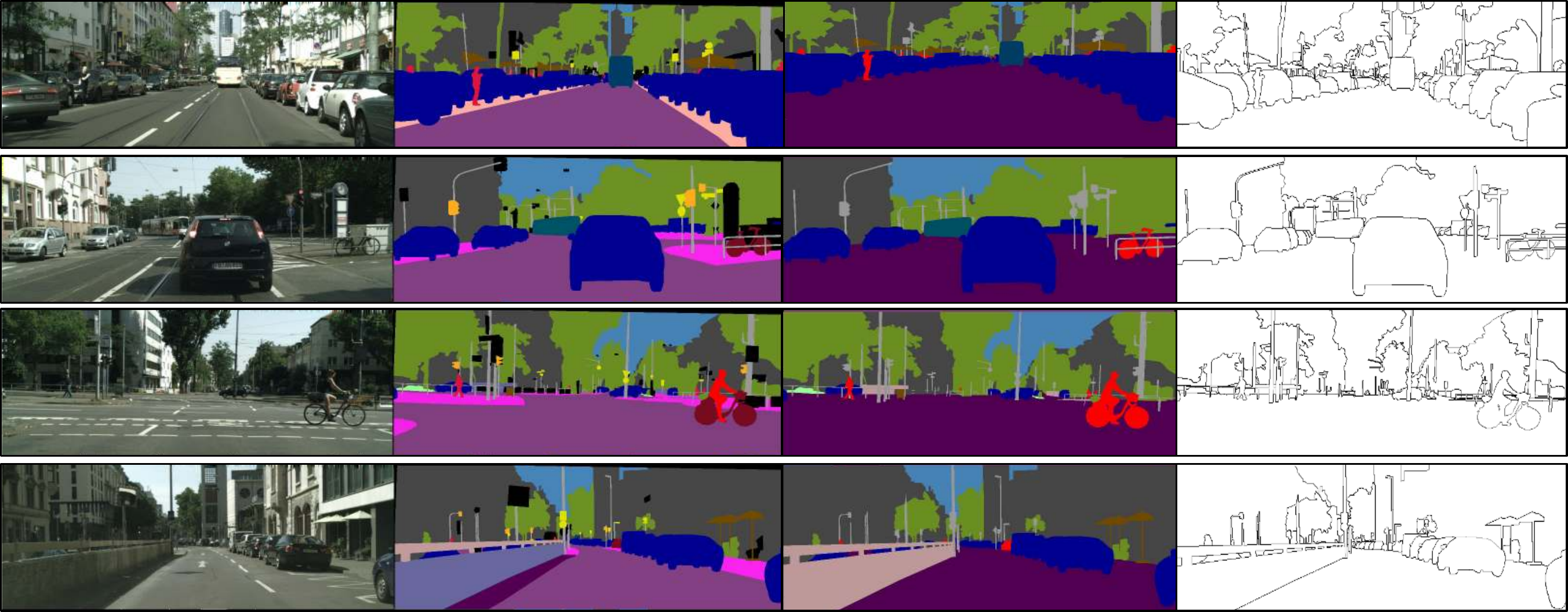}
\caption{The process of geometrical edge ground truth generation. From left to right: RGB images, semantic segmentation ground truth, combined-category segmentation results, geometrical edge ground truth.}
\vspace{\baselineskip}
\label{fig_supp:edge_gt}
\end{figure*}

The edge ground truth of Cityscapes dataset is generated from semantic segmentation ground truth. Some semantic categories share the same 3D surface and are connected in geometrical sense. These geometrically-consistent categories are combined and the geometrical edges are extracted from combined segmentation results. The edges between different instances are preserved in this process. There are four groups of such combining categories as shown in Tab. \ref{tbl:edge_gt}. Examples of the generation of geometrical ground truth are presented in \figref{fig_supp:edge_gt}.
\begin{table}[!htbp]
\centering
\caption{Four groups of semantic categories are combined.}
\label{tbl:edge_gt}
\begin{tabular}{|l|l|}
\hline
Combined Category & Combining Categories                                 \\ \hline
`ground'          & `ground', `road', `sidewalk', `parking'              \\ \hline
`pole'            & `pole', `polegroup', `traffic light', `traffic sign' \\ \hline
`rider'         & `rider', `motorcycle', `bicycle'                     \\ \hline
`wall'            & `wall', `fence', `guard rail'                        \\ \hline
\end{tabular}
\end{table}

\section{Inference between Eqn. 3 and Eqn. 4 / Eqn. 5}
\textbf{From Eqn. 3 to Eqn. 4.} For any two points $x_i$ $x_j$ that lie on the same 3D surface $S$, the surface normal direction should be the same for the two points, which is constrained in Eqn. 4.

\textbf{From Eqn. 3 to Eqn. 5.} For three points $p_i$, $p_j$, $p_k$ that lie on the 3D line, the gradient between any two points should be the same. 

\textbf{From Eqn. 5 to Eqn. 3.} For any three points $p_i$, $p_j$, $p_k$, the gradients between $p_i$, $p_j$ and $p_j$, $p_k$ are the same. Assume the 3D line linking $p_i$, $p_j$ and $p_j$, $p_k$ are represented as:
\begin{align}
a_1x+b_1y+c_1z = 1 \nonumber \\
a_2x+b_2y+c_2z = 1 \nonumber
\end{align}

The gradients are the same for the two lines, thus:
\begin{align}
\frac{a_1}{c_1} = \frac{a_2}{c_2} \nonumber, 
\frac{b_1}{c_1} = \frac{b_2}{c_2} \nonumber
\end{align}
Considering that $p_j$ lies on both lines, thus these two lines are identical. Thus Eqn. 3 and Eqn. 5 are mutually necessary and sufficient conditions.

\section{Example outputs}
\begin{figure*}[!htbp]
\vspace{\baselineskip}
\centering
\includegraphics[width=\textwidth]{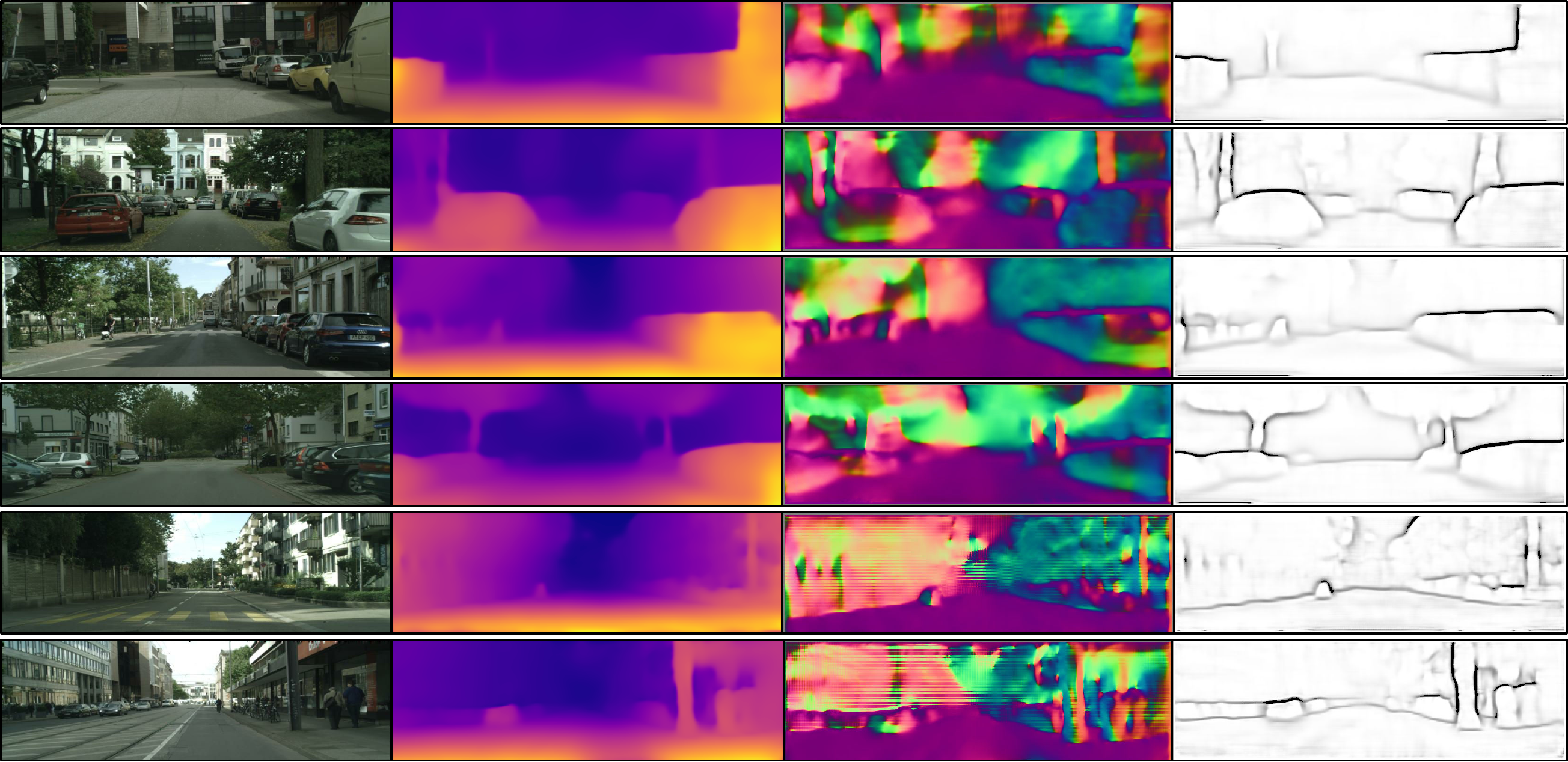}
\caption{Example outputs of LEGO. From left to right: input image, predicted depth, predicted normal, predicted edge.}
\vspace{\baselineskip}
\label{fig_supp:example}
\end{figure*}

LEGO jointly estimates depth, surface normal and geometrical edge. Some example results are shown in \figref{fig_supp:example}.

\section{Comparison with previous methods}
\begin{figure*}
\vspace{\baselineskip}
\centering
\includegraphics[width=\textwidth]{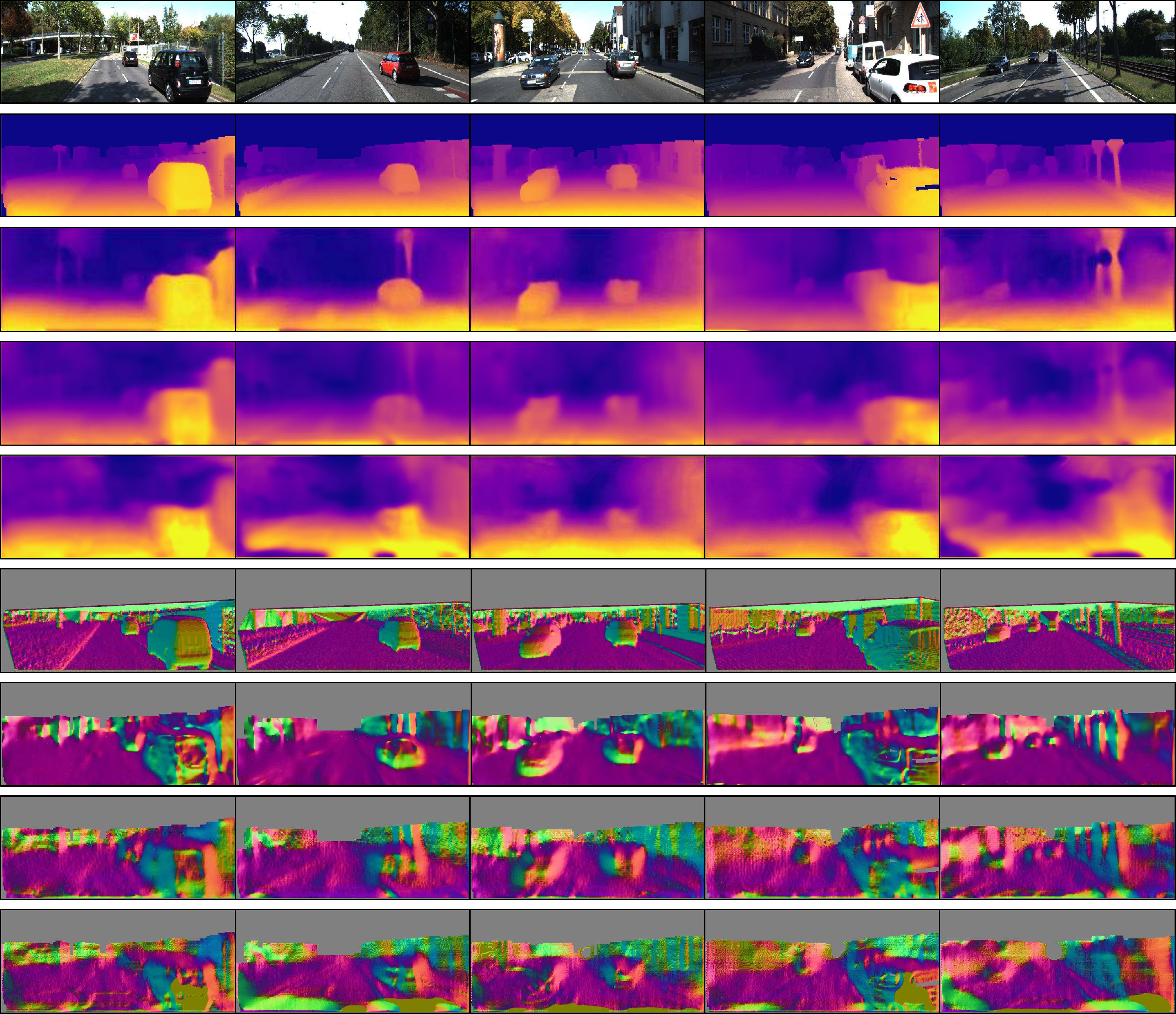}
\caption{Visual results of depth and surface normal by different methods. From top to bottom: input image, depth ground truth, LEGO depth, depth by \cite{yang2018aaai}, depth by \cite{zhou2017unsupervised}, normal ground truth, LEGO normals, normals by \cite{yang2018aaai}, normals by \cite{zhou2017unsupervised}}
\vspace{\baselineskip}
\label{fig_supp:dn_compare}
\end{figure*}

We provide visual comparison with both \cite{zhou2017unsupervised} and \cite{yang2018aaai} in \figref{fig_supp:dn_compare} (see next page). LEGO generates depth and normal results of better structure and preserves aligned object boundaries.

\section{Qualitative results}
Some qualitative results on Cityscapes dataset are shown in the attached video. Cityscapes dataset provides a 30-frame snippet around the key frames. We show 10 snippets in validation set from diverse scenes. The video is available at this link (\url{https://youtu.be/40-GAgdUwI0}).

\clearpage

{\small
\bibliographystyle{ieee}
\bibliography{reference}
}
\end{document}